\pdfoutput=1
\documentclass[10pt, a4paper]{article}
\usepackage{lrec2006}
\makeatletter
\newcommand{\@BIBLABEL}{\@emptybiblabel}
\newcommand{\@emptybiblabel}[1]{}
\makeatother
\usepackage{fancyhdr} 
\usepackage{hyperref}

\title{A Modality Lexicon and its use in Automatic Tagging}

\name{Kathryn Baker$^{\ast\ast\ast}$, Michael Bloodgood$^{\dagger\dagger}$,
 {\bf \large Bonnie J. Dorr$^{\ast}$},  \\ {\bf \large Nathaniel W. Filardo$^{\dagger\dagger}$},  {\bf \large Lori Levin$^{\ast\ast}$},  {\bf \large Christine Piatko$^{\dagger}$}} 

\address{ $^{\ast}$University of Maryland, College Park, MD, bonnie@umiacs.umd.edu \\
          $^{\ast\ast}$Carnegie Mellon University, Pittsburgh, PA, lsl@cs.cmu.edu \\
          $^{\ast\ast\ast}$U.S. Dept. of Defense, Fort Meade, MD, klbake4@tycho.ncsc.mil \\          
          $^{\dagger}$Johns Hopkins Applied Physics Laboratory, Laurel, MD, christine.piatko@jhuapl.edu \\
          $^{\dagger\dagger}$Johns Hopkins HLT Center of Excellence, Baltimore, MD,  bloodgood@jhu.edu, nwf@cs.jhu.edu
}

\abstract{This paper describes our resource-building results for an
  eight-week JHU Human Language Technology Center of Excellence {\it
    Summer Camp for Applied Language Exploration\/} (SCALE-2009) on
  Semantically-Informed Machine Translation.  Specifically, we
  describe the construction of a modality annotation scheme, a
  modality lexicon, and two automated modality taggers that were built
  using the lexicon and annotation scheme.  
  Our annotation scheme is based on identifying three components of
  modality: a trigger, a target and a holder. We describe how our
  modality lexicon was produced semi-automatically, expanding from
  an initial hand-selected list of modality trigger words and phrases.
  The resulting expanded modality lexicon is being made publicly available.
  We demonstrate that one
  tagger---a structure-based tagger---results in precision around 86\%
  (depending on genre) for tagging of a standard LDC data set.  In a
  machine translation application, using the structure-based tagger to
  annotate English modalities on an English-Urdu training corpus
  improved the translation quality score for Urdu by 0.3 Bleu points in
  the face of sparse training data.}

\begin{document}

\thispagestyle{fancy}

\maketitleabstract

\section{Introduction}

This paper describes our resource-building results for an eight-week
JHU Human Language Technology Center of Excellence {\it Summer Camp
for Applied Language Exploration\/} (SCALE-2009) on
Semantically-Informed Machine Translation (SIMT) \cite{SIMT-2009}.
Specifically, we describe the construction of a modality annotation
scheme, a modality lexicon, and two automated modality taggers that
were built using the lexicon and annotation scheme.  Two examples of
modality tagging are shown in Figure~\ref{modality-example}.  Note
that the modality tags are in pairs of triggers and targets. 

In the SIMT paradigm, High Information Value Elements, or HIVEs, are
identified in the English portion of a parallel training corpus and
projected to the source language (in this case, Urdu) during a process
of syntactic alignment, in order to constrain the space of possible
translations.  We explored whether structured annotations of entities
and modalities could improve translation output in the face of sparse
training data and few source language annotations.  Results were
encouraging.  Translation quality, as measured by the Bleu metric
\cite{Papineni:2002}, improved when the training process for the Joshua machine
translation system used in the SCALE workshop \cite{Li:2009} included
modality annotation.

We were particularly interested in identifying modalities because they
can be used to characterize events in a variety of automated analytic
processes.  Modalities can distinguish realized events from unrealized
events, beliefs from certainties, and can distinguish positive and
negative instances of entities and events.  For example, the correct
identification and retention of negation in a particular
language---such as a single instance of the word ``not''---is very
important for a correct representation of events and likewise for
translation.  A major annotation effort for temporal and event
expressions related to the work in this paper is the TimeML
specification language, which has been developed in the context of
reasoning for question answering \cite{SauriVP06}.  TimeML, which
includes modality annotation on events, is the basis for creating the
TimeBank and FactBank corpora \cite{Pustejovsky06,Sauri09}.  In
FactBank, event mentions are marked with their degree of factuality.

The next section defines the theoretical framework we assume in the
creation of our modality lexicon and automatic modality tagger. In
Section~\ref{modality-annotation-scheme}, we described a modality
annotation scheme used by our human annotators.
Section~\ref{English-Modality-Lexicon} describes the creation of a
modality lexicon shared by two types of modality taggers.
Section~\ref{Automatic-Modality-Annotation} describes two different
types of modality taggers: one that is string-based and one that
is structure-based.  Our results and conclusions are then provided.

\section{Modality}
\label{modality-section}
Modality is an extra-propositional component of meaning.  In {\it John
may go to NY\/}, the basic proposition is {\it John go to NY\/} and
the word {\it may\/} indicates modality.  Van der Auwera and
Amman~\cite{VanDerAuweraAmman} define core cases of modality: {\it John
must go to NY\/} (epistemic necessity), {\it John might go to NY\/}
(epistemic possibility), {\it John has to leave now\/} (deontic
necessity) and {\it John may leave now\/} (deontic possibility).  
Many semanticists~\cite{Kratzer,VonFintelIatridou} define modality as
quantification over possible worlds.  {\it John might go\/} means that
there exist some possible worlds in which John goes.  Another view of
modality relates more to a speaker's attitude toward a
proposition~\cite{NirenburgMcShane,McShaneEtAl:2004}.  

\begin{figure}
\begin{center}
\begin{small}
\begin{tabular}{lp{2.7in}}
(1)&{\bf Input:} Americans should know that we can not hand over Dr. Khan to them.\\
   &{\bf Output:} Americans \verb|<|TrigRequire should\verb|>|  \verb|<|TargRequire know\verb|>| that we \verb|<|TrigAble can\verb|>|  \verb|<|TrigNegation not\verb|>| \verb|<|TargNOTAble hand\verb|>|  over Dr. Khan to them.\\  \\
(2)&{\bf Input:} He managed to hold general elections in the year 2002, but he can not be ignorant of the fact that the world at large did not accept these elections.\\
   &{\bf Output:} He \verb|<|TrigSucceed managed\verb|>| to \verb|<|TargSucceed hold\verb|>| general elections in the year 2002, but he \verb|<|TrigAble can\verb|>|  \verb|<|TrigNegation not\verb|>| \verb|<|TargNOTAble be\verb|>| ignorant of the fact that the world at large did \verb|<|TrigNegation not\verb|>| \verb|<|TrigBelief accept\verb|>| these \verb|<|TargBelief elections\verb|>|.
\end{tabular}
\end{small}
\end{center}
\caption{Modality Tagging Examples}
\label{modality-example}
\end{figure}

Modality might be construed broadly to include several types of
attitudes that a speaker might have toward an event or state.  From
the reader or listener's point of view, modality might indicate
factivity, evidentiality, or sentiment.  Factivity is related to
whether an event, state, or proposition happened or didn't happen.  It
distinguishes things that happened from things that are desired,
planned, or probable.  Evidentiality deals with the source of
information and may provide clues to the reliability of the
information.  Did the speaker have first hand knowledge of what he or
she is reporting, or was it hearsay or inferred from indirect
evidence?  Sentiment deals with a speaker's positive or negative
feelings toward an event, state, or proposition.  

Our project was limited to modal words that are related to factivity.
Our focus was on the eight modalities in Figure~\ref{8-modalities},
where P is a proposition and H is the holder (experiencer or cognizer)
of the modality.  Some of the eight factivity-related modalities may
overlap with sentiment or evidentiality.  For example, {\it want\/}
indicates that the proposition it scopes over may not be a fact (it
may just be desired), but it also expresses positive sentiment toward
the proposition it scopes over.  We assume that sentiment and
evidentiality are covered under separate coding schemes, and that
words like {\it want\/} would have two tags, one for sentiment and one
for factivity.

\section{The Modality Annotation Scheme}
\label{modality-annotation-scheme}

The challenge of creating a modality annotation scheme was to deal
with the complex scoping of modalities with each other and with
negation, while at the same time creating a simplified operational
procedure that could be followed by language experts without special
training. 

\subsection{Anatomy of Modality in Sentences} In sentences that
express modality, we identify three components: a trigger, a target,
and a holder.  The trigger is the word or string of words that
expresses modality.  The target is the event, state, or relation that
the modality scopes over.  The holder is the experiencer or cognizer
of the modality.  The trigger can be a word such as {\it should\/},
{\it try\/}, {\it able\/}, {\it likely\/}, or {\it want\/}.  It can
also be a negative element such as {\it not\/} or {\it n't\/}.  Often,
modality is expressed without a lexical trigger.  For a typical
declarative sentence (e.g., {\it John went to NY\/}), the default
modality is strong belief when no lexical trigger is present.
Modality can also be expressed constructionally.  For example,
Requirement can be expressed in Urdu with a dative subject and
infinitive verb followed by the verb {\it parna\/} (to befall).

\begin{figure}
\begin{itemize}
\item {\bf Requirement:} does H require P?
\item {\bf Permissive:} does H allow P?
\item {\bf Success:} does H succeed in P?
\item {\bf Effort:} does H try to do P?
\item {\bf Intention:} does H intend P?
\item {\bf Ability:} can H do P?
\item {\bf Want:} does H want P?
\item {\bf Belief}: with what strength does H believe P?
\end{itemize}
\caption{Eight Modalities Used for Tagging}
\label{8-modalities}
\end{figure}

\subsection{Linguistic Simplifications / Efficient Operationalization}

Six linguistic simplifications were made for the sake of efficient
operationalization of the annotation task. The first linguistic
simplification deals with the scope of modality and negation.  The
first sentence below indicates scope of modality over negation.  The
second indicates scope of negation over modality:
\begin{itemize}
\item He tried not to criticize the president.
\item He didn't try to criticize the president.
\end{itemize}

The interaction of modality with negation is complex, but was
operationalized easily in the menu of thirteen choices shown in
Figure~\ref{13-menu-choices-for-modality}.  First consider the case
where negation scopes over modality.  Four of the thirteen choices are
composites of negation scoping over modality.  For example, the
annotators can choose {\it try\/} or {\it not try\/} as two separate
modalities.  Five modalities (Require, Permit, Want, Firmly Believe,
and Believe) do not have a negated form.  This is because they are
often transparent to negation.  For example, {\it I do not believe
that he left\/} sometimes means the same as {\it I believe he didn't
leave\/}.  Merging the two is obviously a simplification, but it
saves the annotators from having to make a difficult decision.

After the annotator chooses the modality, the scoping of modality over
negation takes place as a second decision.  For example, for the
sentence {\it John tried not to go to NY\/}, the annotator first
identifies {\it go\/} as the target of a modality and then
chooses {\it try\/} as the modality.  Finally, the
annotator chooses {\it false\/} as the polarity of the target.

\begin{figure}
\begin{itemize}
\item H requires [P to be true/false]
\item H permits [P to be true/false]
\item H succeeds in [making P true/false]
\item H does not succeed in [making P true/false]
\item H is trying [to make P true/false]
\item H is not trying [to make P true/false]
\item H intends [to make P true/false]
\item H does not intend [to make P true/false]
\item H is able [to make P true/false]
\item H is not able [to make P true/false]
\item H wants [P to be true/false]
\item H firmly believes [P is true/false]
\item H believes [P may be true/false] 
\end{itemize}
\caption{Thirteen Menu Choices for Modality Annotation}
\label{13-menu-choices-for-modality}
\end{figure}

The second linguistic simplification is related to a duality in
meaning between {\it require\/} and {\it permit\/}.  Not requiring P
to be true is similar in meaning to permitting P to be false.
Thus, annotators were instructed to label {\it not require P to
be true\/} as {\it Permit P to be false\/}.   Conversely, {\it not
Permit P to be true\/} was labeled as {\it Require P to be false\/}.

The third simplification relates to entailments between modalities.
Many words have complex meanings that include components of more than
one modality.  For example, if one managed to do something, one tried
to do it and one probably wanted to do it.  Thus, annotators were
provided a specificity-ordered modality list in
Figure~\ref{13-menu-choices-for-modality}, and were asked to choose the
first applicable modality.  We note that this list corresponds to two
independent ``entailment groupings,'' ordered by specificity:
\begin{itemize}
\item \{{\it requires\/} $\rightarrow$ {\it permits\/}\}
\item \{{\it
  succeeds\/} $\rightarrow$ {\it tries\/} $\rightarrow$ {\it
  intends\/} $\rightarrow$ {\it is able\/} $\rightarrow$ {\it
  wants\/}\}
\end{itemize}
Inside the entailment groupings, the ordering
corresponds to an entailment relation, e.g., {\it succeeds\/} can only
occur if {\it tries\/} has occurred.  Also, the \{{\it requires
  $\rightarrow$ $\ldots$ \/}\} entailment grouping is taken to be more
specific than (ordered before) the \{{\it succeeds $\rightarrow$
  $\ldots$ \/}\} entailment grouping.  Moreover, both entailment
groupings are taken to be more specific than the {\it believes\/},
which is not in an entailment relation with any of the other
modalities.

The fourth simplification, already mentioned above, is that sentences
without an overt trigger word are tagged as Firmly Believes.  This
heuristic works reasonably well for the types of documents we were
working with, although one could imagine genres such as fiction in
which many sentences take place in an alternate possible world
(imagined, conditional, or counterfactual) without explicit marking.

The fifth linguistic simplification is that we did not require
annotators to mark nested modalities.  For a sentence like {\it He
might be able to go to NY\/} the target word {\it go\/} is marked as
ability, but {\it might\/} is not annotated for Belief modality. 
This decision was based on time limits on the annotation task;
there was not enough time for annotators to deal with syntactic
scoping of modalities over other modalities.

Finally, we did not mark the holder H because of the short time frame for
workshop preparation.  We felt that identifying the triggers and
targets would be most beneficial in the context of machine
translation.

\section{The English Modality Lexicon}
\label{English-Modality-Lexicon}

This section describes the creation of a modality lexicon that is
used by the two taggers to be described below in
Section~\ref{Automatic-Modality-Annotation}. Entries in the modality
lexicon consist of: (1) A string of one or more words: for example,
{\it should\/} or {\it have need of\/}.  (2) A part of speech for each
word: the part of speech helps us avoid irrelevant homophones such as
the noun {\it can\/}.  (3) A modality: one of the thirteen modalities
described above.  (4) A head word (or {\it trigger\/}): 
the primary phrasal constituent to
cover cases where an entry is a multi-word unit, e.g., the word {\it
  hope\/} in {\it hope for\/}. (5) One or more subcategorization
codes.

We produced the full English modality lexicon semi-automatically.
First, we used a thesaurus to make a list of modality trigger words
and phrases (about 150 lemmas).  Then we created an inventory of
patterns based on TSurgeon~\cite{Levy:2006}
that show the structural relationship of targets to
triggers for different verb types (further described in
Section~\ref{structured-based-english-modality-tagger} below).  
We defined a mapping between subcategorization codes from Longman's
Dictionary of Contemporary English (LDOCE)~\cite{Procter:1978} and our
TSurgeon patterns.  For example, the LDOCE code T3 corresponds to a
TSurgeon pattern where the modality target is the direct object of
the modality trigger.  We automatically retrieved the LDOCE codes for
the 150 lemmas and used our mapping to assign TSurgeon patterns.  The
150 lemmas were also inflected (four or five forms for each English
verb; singular and plural for nouns).

We note that most intransitive LDOCE codes were not applicable to
modality constructions.  For example, {\it hunger\/} (in the {\it
Want\/} modality class) has a modal reading of ``desire'' when
combined with the preposition {\it for\/} (as in {\it she hungered for
a promotion\/}), but not in its pure intransitive form (e.g., {\it he
hungered all night\/}).  Thus the LDOCE code \verb|I| associated
with the verb {\it hunger\/} was hand-changed to \verb|I-FOR|.
There were 43 such cases.  Once the LDOCE codes
were hand-verified (and modified accordingly), the mapping to 
subcategorization codes was applied.  

The modality lexicon is publicly available at
{ www.umiacs.umd.edu/\~{ }bonnie/ModalityLexicon.txt}.  An
example of an entry is given in Figure~\ref{need-entry},
for the verb {\it need\/}.
\begin{figure}
\begin{center}
\begin{tabular}{lp{2.6in}}
{\bf String:}&Need\\
{\bf Pos:}&VB\\
{\bf Modality:}&Require \\
{\bf Trigger:}&Need\\
{\bf Subcat:}& {\bf V3-passive-basic} --
The government is needed to buy tents.\\
{\bf Subcat:}& {\bf V3-I3-basic} --
The government will need to work continuously
for at least a year.  We will need them to work continuously.\\
{\bf Subcat:}& {\bf T1-monotransitive-for-V3-verbs} --
We need a Sir Sayyed again to maintain this sentiment.\\
{\bf Subcat:}& {\bf T1-passive-for-V3-verb} --
Tents are needed.\\
{\bf Subcat:}& {\bf Modal-auxiliary-basic} --
He need not go.
\end{tabular}
\end{center}
\caption{Modality Lexicon Entry for {\it need\/}}

\label{need-entry}
\end{figure}

\section{Automatic Modality Annotation}
\label{Automatic-Modality-Annotation}

A modality tagger produces text or structured text in which modality
triggers and/or targets are identified.  Automatic identification of
the holders of modalities was beyond the scope of our project because
the holder is often not explicitly stated in the sentence in which the
trigger and target occur.  This section describes two modality
taggers: a string-based English tagger and a structure-based English
tagger. 

\subsection{The string-based English modality tagger}
\label{string-based-english-modality-tagger}

The string-based tagger operates on text that has been tagged with
parts of speech by 
a Collins-style statistical parser~\cite{Miller:1998}.
The tagger marks spans of
words/phrases that exactly match modality trigger words in the
modality lexicon described above, and that exactly match the same
parts of speech. This tagger identifies the target of each modality
using the heuristic of tagging the next non-auxiliary verb to the
right of the trigger.  Spans of words can be tagged multiple times
with different types of triggers and targets.

\subsection{The structure-based English modality tagger}
\label{structured-based-english-modality-tagger}

The structure-based tagger operates on text that has been
parsed~\cite{Miller:1998}.  We used a version of the parser that
produces flattened trees.  In particular the flattener deletes VP
nodes that are immediately dominated by VP or S and NP nodes that are
immediately dominated by PP or NP.  The parsed sentences are processed
by TSurgeon rules.  Each TSurgeon rule consists of a pattern and an
action.  The pattern matches part of a parse tree and the action
alters the parse tree.  More specifically, the pattern finds a
modality trigger word and its target and the action inserts tags such
as {\tt TrigRequire} and {\tt TargRequire} for triggers and targets
for the modality Require.  Figure~\ref{structure-based-tagger-output}
shows output from the structure-based modality tagger. (Note that the
sentence is disfluent: {\it Pakistan which could not reach semi-final,
  in a match against South African team for the fifth position
  Pakistan defeated South Africa by 41 runs.\/}) The example shows
that {\it could\/} is a trigger for the Ability modality and {\it
  not\/} is a trigger for negation.  {\it Reach\/} is a target for
both Ability and Negation, which means that it is in the category of
``H is not able [to make P true/false]'' in our coding scheme.  {\it
  Reach\/} is also a trigger for the Succeed modality and {\it
  semi-final\/} is its target.

\begin{figure}[htb]
\begin{small}
\begin{verbatim}
(TOP
 (S
  (NP 
   (NNP Pakistan) 
   (SBAR (WDT which) 
    (S (MD TrigAble could) 
       (RB TrigNegation not) 
       (VB B TargAble TrigSucceed 
        TargNegation reach) 
       (ADJP 
        (JJ TargSucceed semi-final)) 
        (, ,) 
       (PP (IN in) (DT a) 
           (NN match) (PP (IN against) 
           (ADJP (JJ South) (JJ African))
            (NN team)) 
           (PP (IN for) (DT the)
               (JJ fifth) (NN position))
           (NP (NNP Pakistan)))))) 
  (VB D defeated) 
  (NP (NNP South) (NNP Africa)) 
  (PP (IN by) (CD 41) (NNS runs)) (. .)))
\end{verbatim}
\end{small}
\caption{Sample output from the structure-based modality tagger}
\label{structure-based-tagger-output}
\end{figure}

The TSurgeon patterns are automatically generated from the verb class
codes in the modality lexicon along with a set of templates.  Each
template covers one situation such as the following: the target is the
subject of the trigger; the target is the direct object of the
trigger; the target heads an infinitival complement of the trigger;
the target is a noun modified by an adjectival trigger, etc.  There
are about fifteen templates.  The verb class codes indicate which
templates are applicable for each trigger word.  For example, a
trigger verb in the transitive class may use two target templates, one
in which the trigger is in active voice and the target is a direct
object ({\it need tents\/}) and one in which the trigger is in passive
voice and the target is a subject ({\it tents are needed\/}).

In developing the TSurgeon rules, we first conducted a corpus analysis
for about forty trigger words in order to identify and debug the most
common templates.  We then used LDOCE to assign verb classes to the
remaining verbal triggers in the modality lexicon, and we associated
one or more debugged template with each verb class.  In this way, the
initial corpus work on a limited number of trigger words was
generalized to a longer list of trigger words.  The TSurgeon patterns
will not work with the output of any other parser.  However, the
modality lexicon itself is portable.  If we were to switch parsers, we
would have to write new TSurgeon templates, but the trigger words in
the modality lexicon would still be automatically assigned to
templates based on their verb classes.

\subsection{Agreement of string and structure-based taggers}

We conducted inter-tagger agreement analysis for the string-based and
the structure-based taggers. The Kappa statistic \cite{cohen1960} is
commonly used for measuring agreement, and takes agreement expected
by chance into account. We measured sentence-level agreement between
the string-based and the structure-based taggers for both triggers
and targets.  The average agreement over all the modalities for
triggers was 0.82 and for targets was 0.76.  Since the triggers are
lexicon-based and both taggers used the same lexicon, it is not
surprising the agreement for triggers was relatively high. The disagreements
show where the rule-based tagger is more robust to more complex
parse structure as well as parse errors.  The average target
agreement, at 0.76, was lower than the trigger agreement, which was
also not unexpected. This is because the structure-based tagger's rules for tagging targets are more
complex than the string-based tagger's heuristic for tagging verbs as targets.
The structure-based tagger also sometimes tags nouns as targets, not just verbs.

\section{Results}

To evaluate the effectiveness of our modality tagging, we performed a
manual inspection of the structure-based tagging output. We calculated
precision by examining 249 modality-tagged sentences from the English
side of the NIST 09 MTEval training sentences. We found that 215 tags,
or 86.3\%, were correct.  However, precision of the tags varies with
genre.  In one stretch of 77 sentences from native English newswire
92\% of the tags were correct, whereas the precision may be as low as
83\% for non-native text or text with more complex sentence
structures.  Error analysis revealed the following issues.

First, there were sentences in which a light verb or noun was the
correct syntactic target, but not the correct semantic target.  {\it
Decision\/} would be a better target than {\it taken\/} in {\it The
decision {\bf should} be {\bf taken} on delayed cases on the basis
of merit.\/} Second, since the modality lexicon was used without
respect to word sense, the wrong word sense was tagged.  For example
{\it attacked\/} was part of the lexicon with the intended sense of
{\it try\/} as in {\it attacked the problem\/}, but this did not often
match the word sense for {\it attacked\/} in newswire sentences such
as {\it Sikhs {\bf attacked} a train\/}.  Third, because of the
time-limited nature of our project, we did not write rules to find
triggers and targets in coordinate structures.  Fourth, because of the
flattened parse structures, we could not always identify the head word
of a compound noun correctly and some non-heads were tagged as
targets.  

With respect to recall, the tagger primarily missed special forms of
negation in noun phrases and prepositional phrases: {\it There was
{\bf no} place to seek shelter.\/}; {\it The buildings should be
reconstructed, {\bf not} with RCC, but with the wood and steel
sheets.\/} More complex constructional and phrasal triggers were
also missed: {\it President Pervaiz Musharraf has said that he will
{\bf not rest unless} the process of rehabilitation is completed.\/}
Finally, we discovered some omissions from our modality lexicon: {\it
It is not {\bf possible} in the middle of winter to re-open the
roads.}  Further annotation experiments are planned, which will
be analyzed to close such gaps and update the lexicon as appropriate.

Providing a quantitative measure of recall was beyond the
scope of this project.  At best we could count instances of sentences
containing trigger words that were not tagged.  However, because of
the complexity and subtlety of modality, it would be impractical to
count every clause (such as the {\it not rest unless\/} clause above)
that had a nuance of non-factivity.

We also were able to measure the effect of the modality tagging on the
quality of machine translation output in an Urdu-English machine
translation system, as part of the summer workshop.  A de facto Urdu
modality tagger resulted from identifying the English modality trigger
and target words in a parallel English-Urdu corpus, and then
projecting the trigger and target labels to the corresponding words in
Urdu syntax trees.  English modality annotations alone, as described
in this paper, increased the standard Bleu measure of machine
translation quality from 26.4 to 26.7.  Identifying entities and
modalities in combination increased the score further to 26.9.  It is
future work to also annotate the source language training data
directly with modalities, in order to yield greater translation
quality during alignment and translation. 

In the future, we also plan to investigate practical annotation
concerns (e.g., annotation difficulty) by using multiple
annotators to quantify inter-annotator agreement and also by measuring
the time required for annotation.

\section{Conclusions}

We developed a modality lexicon and a set of automatic taggers, one of
which---the structure-based tagger---results in 86\% precision for
tagging of a standard LDC data set. The modality tagger has been used
to improve machine translation output by imposing semantic constraints
on possible translations in the face of sparse training data.  The
tagger is also an important component of a language-understanding
module for a related project.

\section{Acknowledgments}
We thank Anni Irvine and David Zajic for their help with experiments on an
alternative Urdu modality tagger based on projection and training an HMM-based tagger
derived from Identifinder~\cite{Bikel:1999}. This work is supported, in part, by the
Johns Hopkins Human Language Technology Center of Excellence. Any
opinions, findings, and conclusions or recommendations expressed in
this material are those of the authors and do not necessarily reflect
the views of the sponsor.

\bibliographystyle{lrec2006}
\bibliography{paper}

\end{document}